\documentclass[runningheads]{llncs}
\usepackage[T1]{fontenc}
\usepackage{graphicx,verbatim, tabularx, graphics}
\usepackage{multirow}
\usepackage{booktabs}
\usepackage{url}
\usepackage{cite}
\usepackage{algpseudocode}
\usepackage{tabto}
\usepackage{arydshln}
\usepackage{algorithm}
\usepackage{pifont} 
\usepackage[hidelinks]{hyperref}

\usepackage{amssymb}

\usepackage{amsmath}
\newcommand{\cmark}{\ding{51}}%
\newcommand{\xmark}{\ding{55}}%
\begin{document}
\title{\textsc{EchoingECG}: An Electrocardiogram Cross-Modal Model for Echocardiogram Tasks}
%

\author{Yuan Gao\inst{1, 2, 3, 7} \and
Sangwook Kim\inst{1, 2, 3} \and
Chris McIntosh\inst{1, 2, 3, 4, 5, 6, 7}}
\authorrunning{Y. Gao et al.}

\institute{
Peter Munk Cardiac Centre, University Health Network (UHN), Toronto, Canada \and
Department of Medical Biophysics, UofT, Toronto, Canada \and 
Ted Rogers Centre for Heart Research, UHN, Toronto, Canada \and
Department of Computer Science, University of Toronto (UofT), Toronto, Canada \and
Toronto General Hospital Research Institute, UHN, Toronto, Canada \and
Department of Medical Imaging, UofT, Toronto, Canada \and
Vector Institute, Toronto, Canada \\
\email{\{yuan.gao, sangwook.kim, chris.mcintosh\}@uhn.ca} \\
}
\maketitle              
\begin{abstract}
Electrocardiogram (ECG) is a widely used tool for assessing cardiac function due to its low cost and accessibility. Emergent research shows that ECGs can help make predictions on key outcomes traditionally derived from more complex modalities such as echocardiograms (ECHO), enabling the use of ECGs as a more accessible method to predict broader measurements of cardiac function. ECHO, in particular, are of great importance because they require considerable hospital resources while playing a key role in clinical cardiac assessment. To aid this use case, we introduce \textsc{EchoingECG}, a probabilistic student-teacher model that leverages uncertainty-aware ECG embeddings and ECHO supervision to improve ECG-based cardiac function prediction. Our approach integrates Probabilistic Cross-Modal Embeddings (PCME++), a probabilistic contrastive framework, with ECHO-CLIP, a vision-language pre-trained model trained on ECHO-text pairs, to distill ECHO knowledge into ECG representations. Through experiments and external validation, we showed that \textsc{EchoingECG} outperforms state-of-the-art foundation ECG models in zero-shot, few-shot, and fine-tune settings for ECHO predictions based on ECG. We also highlighted that variance estimation (enabled through our method) enhanced our understanding of model performance by identifying underlying regions of uncertainty within ECGs. The code is available: \url{https://github.com/mcintoshML/EchoingECG}.


\end{abstract}
\def\thickhline{\noalign{\hrule height0.75pt}}

\section{Introduction}
Electrocardiograms (ECGs) are widely used for cardiac assessment due to their accessibility and cost-effectiveness~\cite{attia2019screening}. A growing area of research leverages the use of ECGs to estimate echocardiogram (ECHO) variables to evaluate heart function~\cite{sangha2023detection,yao2021artificial}, which could enhance early disease detection and expand healthcare accessibility, particularly in resource-limited settings. However, replicating ECHO-derived insights from ECGs remains challenging due to \textbf{(1)} complexity of ECG signals and \textbf{(2)} scarcity of ECG data labeled with ECHO outcomes.

\begin{figure}[ht!]
    \centering
    \includegraphics[width=\textwidth]{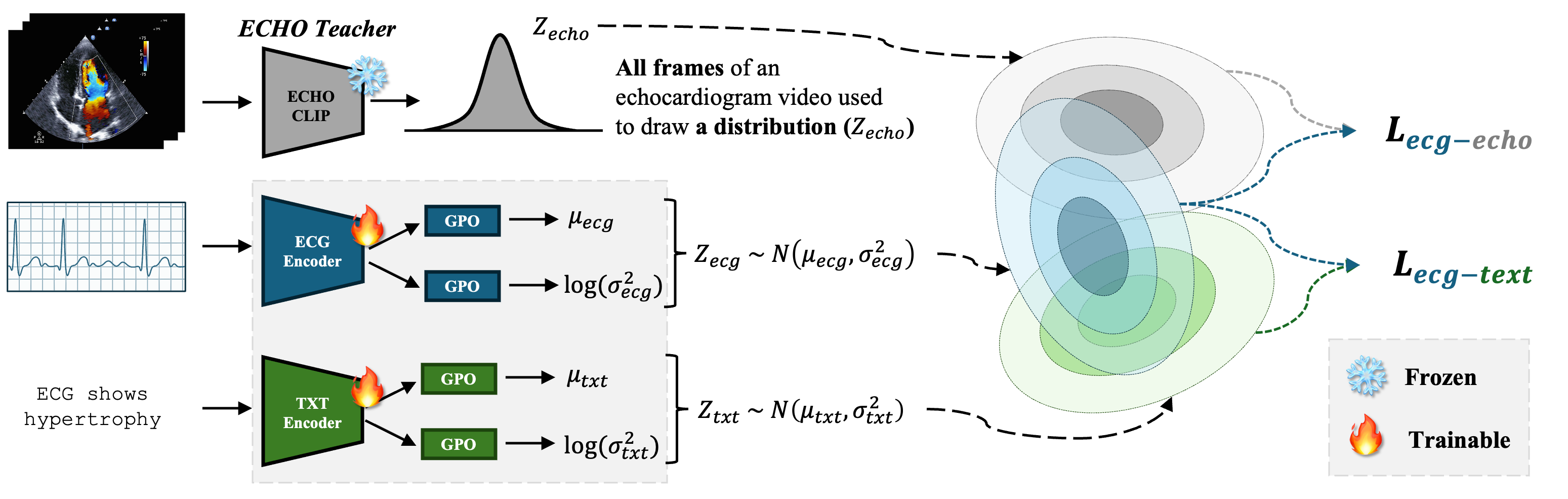}
    \caption{The probabilistic approach employs PCME++ to train the VLPM within a probabilistic space. \textbf{ECHO Teacher (top)}: ECHO-CLIP extracts the distribution’s mean $\mu$ and variance $\log(\sigma^2)$ by utilizing multiple frames from ECHO videos.  \textbf{Student Model (bottom):} Using PCME++, each encoder generates an $l_2$-normalized vector $\mu$ and a vector $\log(\sigma^2)$ to represent a normally distributed random variable. \textsc{EchoingECG} is trained to bind both ECG-Text $L_{ecg-text}$ and ECG-ECHO $L_{ecg-echo}$.} 
    \label{fig:overview}
\end{figure}

For the \textbf{first challenge}, cross-modality contrastive learning offers a promising approach to learning semantically rich embeddings that capture clinically relevant signal patterns~\cite{li2024frozen,mckeen2024ecgfm, gao2024medbind}. Emerging models such as ECG-CLIP, MEDBind~\cite{gao2024medbind}, and others adapt this paradigm, using text descriptions (ECG impressions) for training. These ECG models demonstrate broad applicability in various downstream tasks. However, traditional cross-modal contrastive learning to ECG-ECHO and/or ECG-text integration introduces additional complexities.  ECG signals exhibit subtle variations, with distinct patterns recurring at non-uniform time intervals and manifesting differently across recordings~\cite{berndt1994using,gatzoulis2018signal}. For example, a 120-second ECG is subdivided into many 10-second windows; some windows capture irregular beats, while others do not, yet all windows correspond to the paired report or ECHO. Consequently, injective embeddings in traditional contrastive learning may be limiting as a deterministic embedding space, which means there exists only one-to-one mappings. 

To address this, we propose contrastive cross-modal learning using probabilistic vision-language models (VLPMs), introducing more flexible embeddings. Unlike traditional contrastive learning, probabilistic VLPMs enable many-to-many mappings, better capturing the inherent uncertainty in cross-modality binding. Although we are subject to natural physiological and measurement variations in our framework, different ECG recordings can converge on the same underlying cardiac condition observed in the text or ECHO. This many-to-many mapping reflects the reality that various representations (ECG signals) may all indicate the same cardiac state, thereby enhancing the model’s robustness. We employ Probabilistic Cross-Modal Embeddings (PCME++)~\cite{chun2021probabilisticpcme,chunimprovedpcme++}, which explicitly model uncertainty by learning both the mean and variance of latent representations. By applying the variational information bottleneck principle, PCME++ enables the network to capture uncertainty inherent in paired modalities. This probabilistic structure allows all ECG windows to map to all ECHO frames and/or possible texts, with lower certainty assigned to mappings between less relevant segments. 

To address the \textbf{second challenge}--the need for large-scale annotated ECG-ECHO datasets--recent research has explored using foundation models as teachers to enhance learning in domains with limited labeled data. The Unified Representation of Language, Images, and Point Clouds (ULIP)~\cite{xue2023ulip} exemplifies this, where it leveraged large-scale VLPMs (2D model) to assist training 3D point cloud understanding. ULIP demonstrated that models using a foundation teacher on paired unlabeled data could effectively transfer knowledge to different modalities. We extended this idea by leveraging ECHO-CLIP~\cite{christensen2024visionEchoClip}, a VLPM trained on ECHO-text pairs, as a teacher model to guide ECG representation learning, through \textbf{training on ECHO-ECG pairs}. Our approach distills echo-based cardiac information into ECG representations and, through this alignment, enables ECHO-based training while mitigating large-scale annotated datasets. \par

\textbf{Contributions:} We introduce \textbf{\textsc{EchoingECG}}, a probabilistic student-teacher framework that integrates PCME++ and ECHO-CLIP for enhanced ECG modeling. Our contributions are: \textbf{(1)} the development of a probabilistic multimodal ECG model that captures structural and functional insights from ECHO, \textbf{(2)} a knowledge distillation approach using a student-teacher method that reduces the dependency on large-scale annotated datasets, and \textbf{(3)} experimental evidence showing that probabilistic modeling in \textsc{EchoingECG} enables an uncertainty-aware evaluation of downstream tasks.

\section{Methods and Materials}
\subsection{Probabilistic Contrastive Learning}
\label{ssc:21probab}
Temporal variability is critical in ECGs, as disease manifestation can occur over time. Traditional deterministic contrastive learning methods, InfoNCE~\cite{oord2018representation}, enforce one-to-one mappings, which often struggle to capture the inherent many-to-many relationships in ECGs--where distinct signal patterns can encode similar clinical insights simultaneously. Using a probabilistic embedding space: our framework explicitly models this variance, allowing ECG representations to account for signal dynamics, noise, and artifacts.  To achieve this, we utilized the PCME++ framework~\cite{chunimprovedpcme++}, described below.  

Traditional VLPM binding of ECG embedding $x_{ecg}$ and text embedding $x_{text}$ involves deterministic contrastive learning~\cite{radford2021learning}. Instead, following the probabilistic approach in~\cite{chunimprovedpcme++}, we defined a modality embedding \(x_m\) as a random variable with a multivariate normal distribution: $ x_m \sim \mathcal{N}\left(\mu_m, \sigma_m^2\right)$, where \(\mu_m \in \mathbb{R}^D\) is the mean vector and \(\sigma_m^2 \in \mathbb{R}^D\) denotes the \textbf{diagonal covariance}, as proposed in~\cite{chunimprovedpcme++}. In practice, we optimized $\log(\sigma^2)$, to reduce numerical instability.  PCME++, Figure \ref{fig:overview}, optimizes embeddings to maximize the mutual likelihood of pairs (e.g., ECG signal, $Z_{ecg}$, and its corresponding text, $Z_{text}$), while separating unmatched pairs. This is achieved through PCME++'s Closed-Form Sampled Distance (CSD) metric (Eq. \ref{eq:closedform}), which measures the \textbf{probabilistic distance} between embeddings by combining their $\mu$ and $\sigma^2$.
\begin{equation}
    \label{eq:closedform}
    d\left(Z_{m1},\ Z_{m2}\right)\ =\ ||\ \mu_{m1}\ -\ \mu_{m2}\ ||^2_2\ +\ ||\ \sigma_{m1}^2\ +\ \sigma_{m2}^2||_1\ 
\end{equation}
The properties in Eq. \ref{eq:closedform} are described in detail in~\cite{chunimprovedpcme++}. In short, it involves the first term encouraging distributions between modality pairs to have a proper distance between pairs and the second term encouraging embeddings to reflect certainty through minimal $\sigma^2$ for aligned pairs and larger $\sigma^2$ for ambiguity. 
\subsection{Echo Teacher for ECG-to-ECHO Binding}
\label{ssc:22echotecher}
While ECG-text binding using CSD is applied directly from~\cite{chunimprovedpcme++}, \textsc{EchoingECG} used a \textbf{ECHO Teacher} model that extends learning to strengthen semantic representation through cross-modality binding (ECG-ECHO). We focused on enriching ECG embeddings using cross-modality linking with the existing foundation model: ECHO-CLIP~\cite{christensen2024visionEchoClip}. ECHO-CLIP produces frame-level embeddings for each ECHO video. We denote these embeddings as $e_1,e_2,\ldots,e_n$, where $n$ is the number of frames in the ECHO and, for each $i=1, \dots, n$,  $e_i \in \mathbb{R}^d$  is the embedding of the  $i^\text{th}$ frame with $d$ dimensions. To simplify probabilistic modeling herein, we assumed each frame, $e_i$, of the ECHO as independent. In future work, we aim to explore modeling inter-frame dependencies. 

Thus, we estimated the overall representation of the ECHO video using statistical measures considering the embeddings, $e$, from independent frames as distribution samples. ECHO-CLIP processed each frame; then, we calculated the sample $\mu$ and $\sigma^2$ of the frame embeddings to represent the ECHO video as a probabilistic embedding suitable for PCME++. The $\mu$ embedding is $\frac{1}{n}\sum_{i=1}^{n}e_i$, while $\sigma^2$ embedding: $\sigma^2=\frac{1}{n}\sum_{i=1}^{n}\left(e_i-\mu\right)\odot\left(e_i-\mu\right)$, where $\odot$  denotes element-wise multiplication. By computing these: we modeled the ECHO video as a multivariate normal distribution: $Z_{echo}\sim\mathcal{N}\left(\mu,\sigma^2\right)$, where $\sigma^2$  is a diagonal $\sigma^2$ matrix formed from the $\sigma^2$ vector. Thus, (\ref{eq:closedform}) can be applied to ECG-to-ECHO binding, where we compare $Z_{ecg}$ and $Z_{echo}$ (i.e. $Z_{m1}$ and $Z_{m2}$, respectively).

\subsection{Loss Combination}
\label{ssc:23Loss}
To leverage both ECG-text (Section \ref{ssc:21probab}) and ECG-ECHO learning (i.e., ECHO Teacher proposed in Section \ref{ssc:22echotecher}), we introduce a combined loss function that balances their respective contributions via a hyperparameter $\lambda$. The overall loss used in \textsc{EchoingECG}, $\mathcal{L}_{\mathrm{total}}$, is defined as:
\begin{equation}
\mathcal{L}_{\mathrm{total}}=\lambda\cdot\mathcal{L}_{\mathrm{ecg-text}}+\left(1-\lambda\right)\cdot\mathcal{L}_{\mathrm{ecg-echo}}
\end{equation}
where $\mathcal{L}_{modality-modality}$  represents the probabilistic contrastive loss (Section 2.1), the pairs for modality represent the corresponding ECG-text and ECG-ECHO (student with ECHO Teacher), and $\lambda$ provides a balance for training. 

\subsection{Model Architecture} 
\textsc{EchoingECG} integrates trainable modality-specific encoders to process text and ECG signals.  For the text encoder, we employed BioBERT~\cite{lee2020biobert}, a biomedical variant of BERT~\cite{devlin2018bert}, to capture nuanced medical language relevant to cardiac function. We utilized a 1D-ResNet as the ECG encoder, an efficient framework for extracting features from ECG~\cite{strodthoff2020deep,khan2023ecg}. The primary focus of this research was the training method and highlighting utility of probabilistic embeddings, so we leave architecture changes to future work. Following the text and ECG encoders, two separate linear layers are used--one to produce $\mu$ and the other for $\log{\sigma^2}$. So, an input is mapped to a normal distribution parameterized by $\mu$ and $\log{\sigma^2}$. For ECHO-ECG learning, we used a frozen ECHO-CLIP~\cite{christensen2024visionEchoClip}, as a teacher network, to extract semantic representations of ECHO. ECHO-CLIP processed each frame of an ECHO video separately and $\mu$ and $\log{\sigma^2}$ for each ECHO (Section \ref{ssc:22echotecher}).\par

\subsection{\textsc{EchoingECG} Evaluations}
We tested \textsc{EchoingECG} against ECG foundation models, ECG-CLIP~\cite{gao2024medbind}, ECG-FM (2 versions)~\cite{mckeen2024ecgfm}, and MEDBind~\cite{gao2024medbind}, in ECHO prediction.  Leveraging ECG as an advanced screening tool of traditionally ECHO-derived outcomes could enhance accessibility and early detection of cardiovascular disease ~\cite{al2024prediction,sangha2023detection}. To this end, we tested the predictive capacity on ECHO-derived variables: left ventricular ejection fraction (LVEF), severe left ventricular hypertrophy (SLVH), dilated left ventricular hypertrophy (DVH) using zero-shot (ZS), few-shot (FS), and fine-tune (FT)~\cite{kusunose2025transforming,sangha2023detection,bhave2024deep}. We also assess model performance on traditional ECG-text retrieval. For evaluating probabilistic models, we followed \cite{chunimprovedpcme++}, where $\mu$ embedding was used for prediction, while $\sigma^2$ was used as an uncertainty measure. \par

\noindent\textbf{Implementation Details: }ECG signals were normalized and augmented using standard approaches from~\cite{raghu2022data}. Text inputs consisted of full-length notes, padded to a maximum length of 244 tokens. ECHO videos were processed, following ECHO-CLIP~\cite{christensen2024visionEchoClip}.  All modalities' final $\mu$ and $\log(\sigma^2)$ embedding dimension were 256. The $\lambda$ for our loss (Section ~\ref{ssc:23Loss}) was set to 0.9. Models were trained for 150 epochs, batch size 128, AdamW~\cite{loshchilov2018decoupled_adamw}, weight decay of $1\times{10}^{-1}$, and learning rate of  $4\times{10}^{-4}$. Experiments were conducted using a 48GB L40S GPU.

\section{Experiments and Results}

\subsection{Datasets}
The datasets used are presented in Table \ref{tab:datasets}. 
We pretrained our models using MIMIC-ECG~\cite{gow_mimicecg} and MIMIC-ECHO~\cite{gow_mimicecho}. Patient-level splits were maintained for all MIMIC datasets. \textbf{MIMIC-ECG} dataset contains 10-second recordings of 12-lead ECG signals sampled at 500Hz, which were downsampled to 100 Hz using a low-pass filter~\cite{kher2019signalecg}. We used machine-generated and free-form clinical notes. When free-form text was accessible, it was prioritized; in its absence, machine-generated reports were subsequently utilized for text pairing. \textbf{MIMIC-ECHO} dataset contains ECHOs from MIMIC patients. For our experiments, we used the subset of MIMIC-ECHO, which is comprised of patients who had both an ECG and ECHO during their visit. We derived this subset based on \texttt{subject\_id} and \texttt{admittime} that connects to MIMIC-ECG. \textbf{MIMIC-NOTE} was used to derive ECHO labels~\cite{johnson2023mimic4}. We linked patients who had both ECG and ECHOs with their corresponding discharge note using \texttt{hadm\_id} and extracted ECHO findings following~\cite{goel2023llms} using Llama3.1-instruct 8B~\cite{dubey2024llama}. The MUerte Subita en Insuficiencia Cardiaca \textbf{(MUSIC) ECG} dataset~\cite{martinmusic} comprises vectorcardiograms (VCGs) at 1000Hz. Each recording is associated with ECHO labels of SLVH, DLV, and LVEF. We used the Kors regression transformation~\cite{jaros2019comparison} to convert the VCGs into 12-lead ECGs. The derived ECGs were also downsampled to 100 Hz~\cite{kher2019signalecg}.

\begin{table}[t!]
    \label{tab:dataset}
  \centering
  \caption{Overview of datasets used, tasks, and splits. Modalities column reflects the modality pairs obtained from each dataset. Tasks for each dataset are highlighted, where either pretrain, retrieval, zero-shot (ZS), few-shot (FS), or fine-tune (FT) was conducted with each dataset. $^1$Subset of MIMIC-ECHO with paired ECGs based on \texttt{hadm\_id}. $^2$Only key variables of the ECHO report were provided.}
  
 {
  \def\thickhline{\noalign{\hrule height1.0pt}}
    \begin{tabular}{c;{0.3pt/1.5pt}c;{0.3pt/1.5pt}c;{0.3pt/1.5pt}c;{0.3pt/1.5pt}c;{0.3pt/1.5pt}c}

    \thickhline
    Dataset & Modalities & Task & Train & Valid  & Test \\
    \hline
    MIMIC-ECG & ECG/TXT & Pretrain/Retrieval/ZS/FS & 88,291 & 12,065 & 24,644 \\
    MIMIC-ECHO$^1$ & ECHO/TXT & Pretrain & 814 & 67 & 194 \\
    MUSIC ECG$^2$ & ECG & ZS/FT  & 512 & - & 125 \\  
    \thickhline
    \end{tabular} %
    }
    \label{tab:datasets}
\end{table}

\begin{table}[t!]
\centering
\caption{Zero-shot (ZS) and few-shot (FS), 10\% and 100\%, on LVEF<40\% in MIMIC using ECGs. ECHO labels were derived from ECHO reports. $^*$Uncertainty splits (low/high) based on entropy $>0.5$. $^\dagger$Uncertainty splits in \textsc{EchoingECG} can be calculated directly on median-split of $\sigma^2$ vector, as in~\cite{chunimprovedpcme++}. Balanced accuracy is reported.}
\begin{tabular}{l;{0.3pt/1.5pt}c;{0.3pt/1.5pt}ccc;{0.3pt/1.5pt}ccc;{0.3pt/1.5pt}ccc}
\thickhline
\multirow{3}{*}{Model} & \multirow{3}{*}{Prob?} & \multicolumn{3}{c;{0.3pt/1.5pt}}{\multirow{3}{*}{All}} & \multicolumn{3}{c;{0.3pt/1.5pt}}{Low} & \multicolumn{3}{c}{High}\\
 & & \multicolumn{3}{c;{0.3pt/1.5pt}}{} 
  & \multicolumn{3}{c;{0.3pt/1.5pt}}{Uncertainty} 
  & \multicolumn{3}{c}{Uncertainty} \\ 
& & ZS & 10\% & 100\% & ZS & 10\% & 100\%  & ZS & 10\% & 100\% \\  

\hline
ECG-CLIP~\cite{gao2024medbind} & \xmark & 56.3 & 64.5 & \underline{77.1} & - & \underline{66.2}$^*$ & \underline{77.4}$^*$ & - & 62.8$^*$ & \textbf{76.8}$^*$ \\
MEDBind~\cite{gao2024medbind} & \xmark& \underline{57.5} & \underline{66.9} & 76.2 & - & 61.0$^*$ & 75.7$^*$ & - & \textbf{72.8}$^*$ & \underline{76.7}$^*$ \\
ECG-FM$_{M}$~\cite{mckeen2024ecgfm} & \xmark& - & 55.0 & 66.8 & - & 58.4$^*$ & 69.1$^*$ & - & 51.6$^*$ & 64.5$^*$ \\
ECG-FM$_{P}$~\cite{mckeen2024ecgfm} & \xmark& - & 61.5 & 70.7 & - & 64.2$^*$ & 72.4$^*$ & - & 58.8$^*$ & 69.0$^*$ \\
\hline
\textsc{EchoingECG} (Ours) & \cmark & \textbf{63.2} & \textbf{69.3} & \textbf{78.9} & 68.3$^\dagger$ & \textbf{74.3}$^\dagger$ &\textbf{84.4}$^\dagger$ & 58.1$^\dagger$ & \underline{64.2}$^\dagger$  & 73.4$^\dagger$ \\
\thickhline
\end{tabular}


\label{tab:mimic_lvef}
\end{table}

\subsection{Predicting ECHO-derived Labels using ECGs}
\textbf{MIMIC Results}: We benchmarked \textsc{EchoingECG} on ECHO-related tasks within MIMIC, focusing on ZS and FS for predicting LVEF<40\% using only ECG in Table \ref{tab:mimic_lvef}. In ZS, we computed cosine distance between text prompts and ECG~\cite{wang2022medclip}, while FS used embeddings from frozen ECG encoders with linear probing~\cite{girdhar2023imagebind}. Note, \textbf{ECG-FM} is unimodal. Thus, all \textbf{ECG-FM} ZS results are omitted. Additionally, Table \ref{tab:mimic_lvef} also highlights the impact of data uncertainty on predicting LVEF<40\%, quantified via baseline methods (deterministic) and our method (probabilistic). \textbf{For deterministic:} we can quantify uncertainty in FS results by entropy $H[x]=p(x)\log_2{p(x)}+(1-p(x))\log_2{(1-p(x))}$~\cite{abdar2021review}, where $x$ represents the input and $p(x)$ is the softmax output from linear probing. For simplicity, we considered an entropy above 0.5 as high, while below was low uncertainty. \textbf{For probabilistic (ours):} we can split data with the $\sigma^2$ vector~\cite{chunimprovedpcme++} that \textsc{EchoingECG} calculates for each embedding, where, a median split of the test set into high- and low-$\sigma^2$. \textsc{EchoingECG} predicts $Z_m\sim\mathcal{N}\left(\mu_m, \sigma_m^2\right)$; thus, the predicted $\sigma^2$ reflects uncertainty as it measures the spread of $Z_m$.

In both ZS and FS, \textsc{EchoingECG} outperformed foundation ECG models, highlighting the value of our method in improving ECHO-related task performance. Additionally, no clear relationship can be observed when splitting deterministic methods using entropy. In contrast, splitting via $\sigma^2$ highlights that \textsc{EchoingECG} can appropriately split these subsets. From Table \ref{tab:mimic_lvef}, we observed that \textit{lower} uncertainty (via $\sigma^2$ embedding) corresponded to better performance and vice versa--showing evidence of correct uncertainty estimation behavior.
\begin{table}[t]
\centering
\caption{Comparison with state-of-the-art for zero-shot (ZS) and fine-tune (FT) (100\%) on ECHO-related classification in MUSIC dataset. MUSIC ECGs have longer time windows (20-minutes) vs. traditional ECGs (10-seconds). We report the start result, as it performed best on baseline models. Balanced accuracy is reported. }
\label{tab:model_performance}
\begin{tabular}{l;{0.3pt/1.5pt}cc;{0.3pt/1.5pt}cc;{0.3pt/1.5pt}cc;{0.3pt/1.5pt}cc|cc}
\thickhline
\multirow{2}{*}{Model} & \multicolumn{2}{c;{0.3pt/1.5pt}}{LVEF<40\%} & \multicolumn{2}{c;{0.3pt/1.5pt}}{SLVH} & \multicolumn{2}{c;{0.3pt/1.5pt}}{DLV} & \multicolumn{2}{c|}{SLVH+DLV}  & \multicolumn{2}{c}{Ovr. Rank} \\
& ZS &     FT &ZS &     FT &      ZS &     FT &  ZS &     FT  & ZS & FT \\
\hline
ECG-CLIP (start)~\cite{gao2024medbind} & 64.1 & 58.8 & 58.9 & 59.4 & 60.4 & 58.7 & 61.3 & 84.9 & 5 & 7
 \\
MEDBind (start)~\cite{gao2024medbind} & 64.2 & 60.2 & 56.3 & 62.0 & 61.1 & 61.5 & \underline{62.6} & 85.5 & 4 & 5
 \\
ECG-FM$_{M}$ (start)~\cite{mckeen2024ecgfm} &  - &  63.2 & - &   76.0 &- &  \textbf{69.6} &   - &  87.2 & - & 3 \\
ECG-FM$_{P}$ (start)~\cite{mckeen2024ecgfm} &  - &  64.8 & - & 60.8 &- &  62.4 &- &  74.4 & - & 5 \\
\hline
  \textsc{EchoingECG} (start) & \underline{66.4} &   \textbf{66.0} &  76.0 &  \underline{77.5} & 64.0 &  57.6 &  62.4 &   \underline{89.0} & \underline{2} & \underline{2} \\
 \textsc{EchoingECG} (High $\sigma^2$) & 65.0 &    60.0 &    \underline{77.4} &  76.8  &   \underline{65.6} &  51.2 &  60.6 &   \underline{89.0}  & 3 & 4 \\
  \textsc{EchoingECG} (Low $\sigma^2$) &  \textbf{68.0} &  \underline{65.6} &    \textbf{79.2} &  \textbf{77.6} &   \textbf{70.4} &  \underline{63.2} &  \textbf{66.4} &  \textbf{91.4} & \textbf{1} & \textbf{1}\\
\thickhline
\end{tabular}
\end{table}
\begin{figure}[t!]
    \centering
    \includegraphics[width=0.94\textwidth]{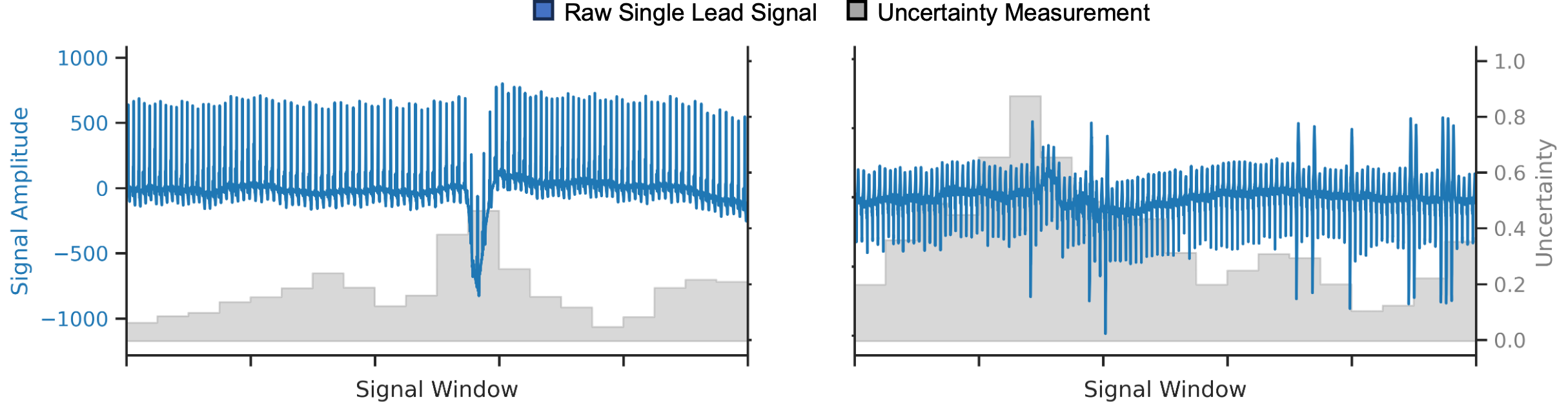}
    \caption{Uncertainty in MUSIC dataset using \textsc{EchoingECG}. 10-second sliding window of ECG Lead I in uncertainty in the ECGs are visualized. Areas of signal instability are associated with \textbf{higher $\sigma^2$}, suggesting that $\sigma^2$ captures the uncertainty.} 
    \label{fig:examples}
\end{figure}

\textbf{External MUSIC Results:} We evaluated \textsc{EchoingECG} performance on external MUSIC dataset for both ZS and FT tasks--focusing on a use-case in handling more general ECGs (i.e. ECGs longer than the 10-second durations seen in training), in Table \ref{tab:model_performance}. Traditional methods typically process short windows to tackle longer ECGs—often extracted from the beginning or a random offset—to handle longer recordings~\cite{mehari2022advancing}. We observed that using the start windows worked best for the baseline models reported herein. In contrast, \textsc{EchoingECG} leverages uncertainty estimation: higher values of $\sigma^2$ indicate greater dispersion, while lower uncertainty corresponds to more consistent signal characteristics. This uncertainty metric guides the selection of the most informative window, enabling our approach to outperform baseline models across multiple tasks consistently. Figure \ref{fig:examples} visualizes this relationship, showing that noisy ECG segments exhibit higher uncertainty, whereas stable signals correspond to lower uncertainty.
\begin{table}[ht!]
\centering
\caption{Top-k retrieval (100\%) on \textbf{text-to-ECG retrieval}. We report the top-1 and top-10 recall (R@k). $^*$Low vs. high uncertainty based on median-split of $\sigma^2$ embedding.
}

\begin{tabular}{lc;{0.3pt/1.5pt}cc;{0.3pt/1.5pt}cc;{0.3pt/1.5pt}cc}
\thickhline
\multirow{2}{*}{Model} & \multirow{2}{*}{Trained Modalities}   & \multicolumn{2}{c;{0.3pt/1.5pt}}{\multirow{2}{*}{All}} 
  & \multicolumn{2}{c;{0.3pt/1.5pt}}{Low} 
  & \multicolumn{2}{c}{High} \\ 
&  & \multicolumn{2}{c;{0.3pt/1.5pt}}{} 
  & \multicolumn{2}{c;{0.3pt/1.5pt}}{Uncertainty$^*$} 
  & \multicolumn{2}{c}{Uncertainty$^*$} \\ 

& & R@1 & R@10 & R@1 & R@10 & R@1 & R@10 \\  
\hline
ECG-CLIP\cite{gao2024medbind} & ECG+TXT & 50.2 & \underline{93.9} & - & - & - & - \\
MEDBind\cite{gao2024medbind} & ECG+TXT+CXR & \underline{53.6} & \textbf{94.5} & - & - & - & - \\
\hline
\textsc{\textsc{EchoingECG}} (Ours) & ECG+TXT+ECHO & \textbf{67.0} & \textbf{94.5} & 76.8 & 98.5 & 45.9 & 72.7 \\
\thickhline
\end{tabular}

\label{tab:topk}
\end{table}

\subsection{Traditional Top-K Text-ECG Retrieval}
We further tested Text-ECG retrieval using recall at top-K~\cite{wang2022medclip,gao2024medbind} to measure alignment between ECGs and text (Table \ref{tab:topk}), \textsc{EchoingECG} remained competitive with other models--showing that adding our combined loss did not compromise ECG-text alignment. We also observed that ECGs with low uncertainty had better retrieval (Table \ref{tab:topk}), reinforcing the learned uncertainty in our approach. 

\subsection{Ablation} 
We conducted an ablation study comparing the performance of InfoNCE~\cite{radford2021learning}, PCME++~\cite{chunimprovedpcme++}, and the proposed additional ECHO Teacher loss in \textsc{EchoingECG}. As shown in Table \ref{tab:ablation}, InfoNCE loss performed poorly in all metrics, highlighting the limitation of deterministic methods in capturing the temporal and structural complexity of multimodal ECG models, where many-to-many relationships may exist. This observation is supported by the PCME++ loss results, which marked improvements and demonstrated a potential improvement in training due to the probabilistic nature of the loss. Finally, our proposed addition achieved the highest performance across all ECHO-based classification tasks. This underscores the guiding model with the ECHO Teacher and how our method could support using ECGs to emulate the structural and functional insights derived from ECHO.
\begin{table}[t!]
\centering
\caption{Ablation on various tasks. For classification (ACC), we highlighted \textbf{ZS} results. \textbf{For all tasks: } all samples were considered, not just low-uncertainty, for fair comparisons. For MUSIC, start window was used for all models.}
\label{tab:ablation}
\begin{tabular}
{l;{0.3pt/1.5pt}c;{0.3pt/1.5pt}c;{0.3pt/1.5pt}c;{0.3pt/1.5pt}c}
\thickhline
\multirow{2}{*}{Methods} & \multirow{2}{*}{Prob?} & MIMIC-ECHO & MUSIC & MIMIC-ECG \\ 
& & LVEF (ACC) & SLVH+DLV (ACC)  & Retrieval (R@1) \\
 \hline
InfoNCE Loss & \xmark & 56.8 & 61.0 & 53.2 \\ 
+ ECHO Teacher & \xmark & 61.2 & \underline{62.1} & 52.9 \\ 
\hline
PCME++ Loss & \cmark & \underline{62.0} & 61.7 & \textbf{67.9} \\ 
+ ECHO Teacher (Ours) & \cmark & \textbf{63.2} & \textbf{62.4} & \underline{67.0}\\ 
\thickhline
\end{tabular}
\end{table}

\section{Conclusion}
This work underscores the importance of addressing the many-to-many mapping relationships inherent in ECG, where deterministic approaches often fail to capture ambiguity. Our proposed \textsc{EchoingECG} model demonstrated improved robustness across ECHO-related tasks by leveraging probabilistic modeling and a foundational model as a teacher (i.e. ECHO-CLIP). Finally, we enhanced interpretability of model predictions through probabilistic uncertainty estimation.

%
%
\bibliographystyle{splncs04}

\end{document}